%% file: 0_main.tex
\documentclass[runningheads]{llncs}
\usepackage{graphicx}
\usepackage{comment}
\usepackage{amsmath,amssymb} 
\usepackage{color}
\usepackage{xcolor}

\usepackage{epsfig}
\usepackage{xspace}
\usepackage{multirow}
\usepackage{microtype} 
\usepackage{cite}
\definecolor{citecolor}{HTML}{0071bc}
\definecolor{linkcolor}{HTML}{e802af}
\definecolor{urlcolor}{HTML}{00a600}
\usepackage[pagebackref=false,breaklinks=true,colorlinks,citecolor=citecolor,linkcolor=linkcolor, urlcolor=urlcolor, bookmarks=false]{hyperref}

\makeatletter
\DeclareRobustCommand\onedot{\futurelet\@let@token\@onedot}
\def\@onedot{\ifx\@let@token.\else.\null\fi\xspace}
\def\eg{\emph{e.g}\onedot}

\def\vs{\emph{vs}\onedot}
 
\def\etal{\emph{et al}\onedot}
\makeatother

\newcommand\dtdd{\textsc{DTD$^2$}\xspace}
\newcommand\dtd{\textsc{DTD}\xspace}
\newcommand\para[1]{\noindent\textbf{#1.}}

\begin{document}
\pagestyle{headings}
\mainmatter
\def\ECCVSubNumber{384}  

\title{Describing Textures using Natural Language} 

\authorrunning{C. Wu, M. Timm, S. Maji}

\newcommand{\orcid}[1]{\,\href{https://orcid.org/#1}{\protect\includegraphics[width=8pt]{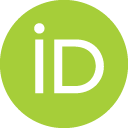}}}
\author{Chenyun Wu\orcid{0000-0001-6542-9279} \and
Mikayla Timm\orcid{0000-0001-5443-243X} \and
Subhransu Maji\orcid{0000-0002-3869-9334}}

\institute{University of Massachusetts Amherst \\ \email{\{chenyun, mtimm, smaji\}@cs.umass.edu}}

\maketitle

\begin{abstract}
Textures in natural images can be characterized by color, shape,
periodicity of elements within them, and other attributes that can be described using
natural language.
In this paper, we study the problem of describing visual attributes of texture on a novel dataset containing rich descriptions of textures, and conduct a systematic study of current generative and discriminative
models for grounding language to images on this dataset.
We find that while these models capture some properties of texture, they
fail to capture several compositional properties, such as the colors of dots.
We provide critical analysis of existing models by generating synthetic but realistic textures
with different descriptions. 
Our dataset also allows us to train interpretable models and generate
language-based explanations of what discriminative features are learned by deep networks for fine-grained
categorization where texture plays a key role.
We present visualizations of several fine-grained domains and show
that texture attributes learned on our dataset offer improvements over
expert-designed attributes on the Caltech-UCSD Birds dataset.
\end{abstract}

\input{1_intro.tex}

\input{2_related.tex}

\input{3_data.tex}

\input{4_method.tex}
\input{5_experiment.tex}

\input{6_analysis.tex}
\input{7_application.tex}

\input{8_conclusion.tex}

\clearpage
%
%
\bibliographystyle{splncs04}
\bibliography{0_main}

\clearpage

\input{supp/supp.tex}

\end{document}

%% file: 1_intro.tex
\section{Introduction}\label{sec:intro}
Texture is ubiquitous and provides useful cues for a wide range of
visual recognition tasks.
We rely on texture for estimating material properties of surfaces,
for discriminating objects with a similar shape, for
generating realistic imagery in computer graphics applications, and so on.
Texture is localized and can be more easily modeled than shape that is
affected by pose, viewpoint, or occlusion.
The effectiveness of texture for perceptual tasks is also mimicked by
deep networks trained on current computer vision datasets that have
been shown to rely significantly on texture for
discrimination
(\eg,~\cite{geirhos2018imagenet,cimpoi2015deep,brendel2019approximating,hosseini18assessing}).

\begin{figure}[t]
\centering
\includegraphics[width=\linewidth]{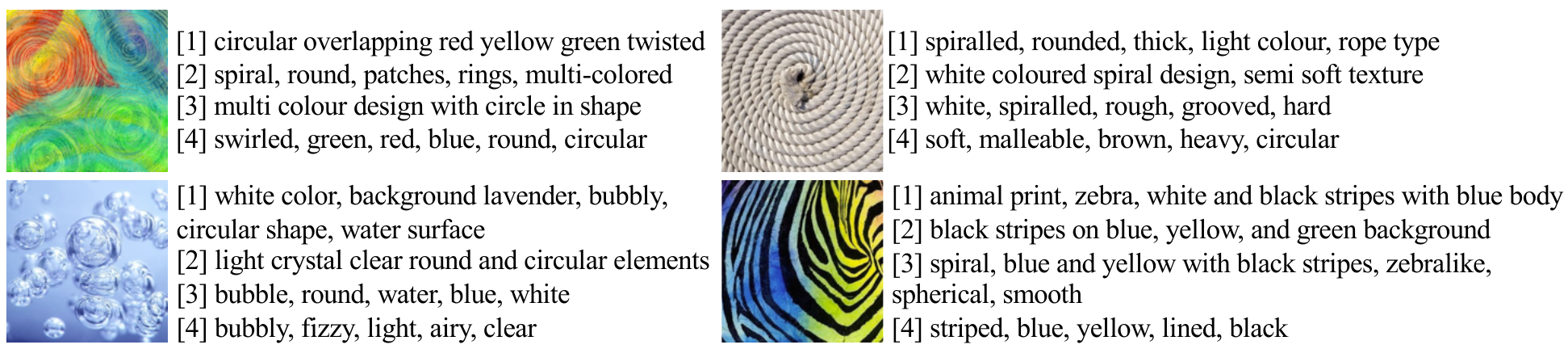}

\caption{We introduce the Describable Textures in Detail Dataset 
  (\dtdd) consisting of texture images from DTD~\cite{cimpoi14describing} with natural language
  descriptions, which provide rich and fine-grained supervision for
  various aspects of texture such as color compositions, shapes, and
  materials. More examples are in the Supplementary material.
  }
\label{fig:data}
\end{figure}

While there has been significant work in the last few decades on
visual representations of texture, limited work has been done on
describing detailed properties of textures using natural language.
The ability to describe texture in rich detail can enable applications
on domains such as fashion and graphics, as well as to interpret
discriminative attributes of visual categories within a fine-grained taxonomy (\eg, species of
birds and flowers) where texture cues play a key role.
However, existing datasets of texture (\eg,~\cite{cimpoi14describing,bell2015material}) are limited to a
few binary attributes that describe patterns or materials, and do not
describe detailed properties using the compositional nature of language (\eg, descriptions of the
color and shape of texture elements).
At the same time, existing datasets of language and
vision~\cite{VQA, lin2014microsoft, sharma2018conceptual,  plummer2015flickr30k, kazemzadeh2014referitgame, mao2016generation, yu2016modeling} primarily focus on objects and their
relations with a limited treatment of textures (Section~\ref{sec:related}).
Addressing this gap in the literature, we introduce a new dataset containing natural
language descriptions of textures called the Describable Textures in Detail Dataset 
(DTD$^2$).
It contains several manually annotated descriptions of each image from the Describable
Texture Dataset (DTD)~\cite{cimpoi14describing}.
As seen in Figure~\ref{fig:data}, these contain descriptions of colors of the
structural elements within the texture (\eg ``circles'' and ``stripes''), their shape, and other
high-level perceptual properties of the texture (\eg ``soft'' and ``protruding'').
The resulting vocabulary vastly extends the 47 attributes present in
the original DTD dataset (Section~\ref{sec:dataset}).

We argue that the domain of texture is rich and poses many
challenges for compositional language modeling that are present in
existing language and vision datasets describing objects and scenes.
For example, to estimate the color of dots in a dotted texture the
model must learn to associate the color to the dots and not to the
background.
Yet the domain of texture is simple enough that it allows us to
analyze the robustness and generalization of existing
vision and language models by synthetically generating variations of a
texture.
We conduct a systematic study of existing visual representations of
texture, models of language, and methods for matching the two domains
on this dataset (Sections~\ref{sec:method}, \ref{sec:exp} and \ref{sec:analysis}).
We find that adopting pre-trained language models significantly improve
generalization.
However, an analysis on synthetically generated variations of each
texture by varying one attribute at a time (\eg, foreground color and
shape) shows that the representations fail to capture detailed properties.

We also present two novel applications of our dataset (Section~\ref{sec:app}).
First, we visualize what discriminative texture properties are learned by
existing deep networks for fine-grained classification on natural domains such as
birds, flowers, and butterflies.
To this end we generate ``maximal images'' for each category by
``inverting'' a texture-based classifier~\cite{lin2018bilinear} and describe
these images using captioning models trained on DTD$^2$. 
We find that the resulting explanations tend to be well aligned with the
discriminative attributes of each category (\eg, ``Tiger
Lily'' flower is ``black, red, white, and dotted'' as seen in Figure~\ref{fig:fgvc}-middle).
We also show that models trained on \dtdd offer improvements over
expert-designed binary attributes on the Caltech-UCSD Birds dataset~\cite{WahCUB_200_2011}.
This complements the capabilities of
existing datasets for explainable AI on these domains that focus on
shapes, parts, and their attributes such as color.
Texture provides a domain-independent, albeit incomplete way of
describing interpretable discriminative properties for several domains.

\noindent
In summary, our contributions are:
\begin{itemize}
  \item A novel dataset of
    texture descriptions (Section~\ref{sec:dataset}).
  \item An evaluation of existing models of grounding
    natural language to texture (Section~\ref{sec:method} and
    \ref{sec:exp}).
  \item A critical analysis of these models
    using synthetic, but realistic variations of textures with their
    descriptions (Section~\ref{sec:analysis}).
  \item Application of our models for describing
    discriminative texture attributes and building interpretable models on  
    fine-grained domains (Section~\ref{sec:app}).
\end{itemize}
Our dataset and code are at: \url{https://people.cs.umass.edu/~chenyun/texture}.

%% file: 2_related.tex
\section{Related Work}
\label{sec:related}

\para{Language and texture}
Describing textures using language has a long history.
Early works~\cite{amadasun1989textural,tamura1978textural,bajcsy1973computer}
showed that textures can be categorized along a few
semantic axes such as ``coarseness'', ``contrast'', ``complexity'' and
``stochasticity''.
Bhusan~\etal~\cite{bhushan1997texture} systematically identified words
in English that correspond to visual textures and analyzed their
relationship to perceptual attributes of textures.
This was the basis of the Describable Texture
Dataset~(\dtd)~\cite{cimpoi14describing} which consolidated a list of 47
texture attributes along with images downloaded from the Internet.
The dataset captures attributes such as ``dotted'', ``chequered'', and ``honeycombed''.
However, it does not capture properties such as the color of
the structural elements (``red and green dots''), or the attributes
that describe the background color.
Our goal is to model the rich space of texture attributes in a compositional manner beyond these attributes. 

\para{Datasets of images and text}
The vision and language community has put significant efforts into
building large-scale datasets.
Image captioning datasets such as MS-COCO~\cite{lin2014microsoft},
Flickr30K~\cite{young2014image} and Conceptual
Captions~\cite{sharma2018conceptual} contain sentences describing the
general content of images.
The Visual Question Answering dataset~\cite{VQA} provides language
question and answer pairs for each image, which requires more detailed
understanding of the image content.
In visual grounding datasets such as
RefClef~\cite{kazemzadeh2014referitgame},
RefCOCO~\cite{mao2016generation, yu2016modeling} and Flickr30K
Entities~\cite{plummer2015flickr30k}, detailed descriptions of the
target object instances are annotated to distinguish them from other
objects.
However, these tasks focus on recognizing object categories and
descriptions of pose, viewpoint, and their relationships to
other objects, and have a limited treatment of attributes related to
texture.

\para{Texture representations}
Representations based on orderless aggregations of local features
originally developed for texture has had an significant influence on early 
computer vision (\eg,
``Textons''~\cite{leung2001representing},``Bag-of-Visual-Words''~\cite{csurka2004visual},
higher-order statistics~\cite{portilla2000parametric}, and Fisher
vector~\cite{perronnin2010large,sanchez2013image}).
Recent works~(\eg,~\cite{cimpoi2015deep,lin2015bilinear,arandjelovic2016netvlad})
have shown that combining texture representations with deep networks
lead to better generalization on scene understanding and fine-grained categorization tasks.
Even without explicit modeling, deep networks are capable of
modeling texture through convolution, pooling, and non-linear encoding
layers~\cite{gatys2016image}.
Indeed, several works have shown that deep networks trained on
existing datasets tend to rely more on texture than shape for
classification~\cite{geirhos2018imagenet,brendel2019approximating,hosseini18assessing,lin2016visualizing}.
This motivates the need to develop techniques to describe texture
properties using natural language as a way to explain the behavior of deep
networks in an interpretable manner.

\para{Methods for vision and language}
There is a significant literature on techniques for various language and vision tasks.
The Show-and-Tell~\cite{vinyals2015show} model was an early
deep neural net based approach for captioning images that
combined the convolutional image encoder followed by an
LSTM~\cite{hochreiter1997long} language decoder.
Techniques for VQA are based on a
joint encoding of the image and the question to retrieve or generate
an answer~\cite{tan2019lxmert,yu2019deep, kim2018bilinear}.
For visual grounding, where the goal is to identify a region in the image given a ``referring expression'', a common approach is to learn
a metric over expressions and regions~\cite{yu2018mattnet, liu2017referring,plummer2018conditional}.
The basic architectures for these tasks have been improved in a number
of ways such as by incorporating attention
mechanisms~\cite{liu2019improving, anderson2018bottom,wang2019neighbourhood, yu2019deep, gao2019dynamic, kim2018bilinear} and improved
language models~\cite{devlin2018bert,schuster1997bidirectional}.
To model the relation between texture images and their descriptions we investigate a discriminative approach, a metric-learning based approach, and a generative modeling based approach~\cite{xu2015show} on our dataset.

%% file: 3_data.tex
\section{Dataset and Tasks}
\label{sec:dataset}

We begin by describing how we collected \dtdd in
Section~\ref{sec:data-collection}, followed by the tasks and
evaluation metrics in Section~\ref{sec:data-tasks}.
\dtdd contains multiple crowdsourced descriptions for each image
in \dtd.
Each image $I$ contains $k$ descriptions 
$\mathbf{S} = \left\{S_1, S_2, \ldots, S_k\right\}$ from $k$ different
annotators who are asked to describe the texture presented in the image.
Instead of providing a grammatically coherent sentence, we found that it more
effective for them to list a set of properties separated by
commas.
Thus each description $S$ can be interpreted as a set of phrases
$\left\{P_1, P_2, \ldots, P_n\right\}$.
Figure~\ref{fig:data} shows some examples of the collected data.
We found that the ordering of phrases in a description is somewhat arbitrary,
which motivates this annotation structure.
Figure~\ref{fig:stats} shows the overall dataset statistics.
\dtdd contains 5,369 images and 24,697 descriptions.
We split the images into $60\%$ training, $15\%$ validation, and
$25\%$ test.
Below we describe details of the dataset collection pipeline and
tasks.

\subsection{Dataset Collection}\label{sec:data-collection}

\para{Annotation}
We present each \dtd image and its corresponding texture category
to $5$ different annotators on Amazon Mechanical Turk, asking them to
describe the texture using natural language with at least $5$ words.
Describable aspects of each image include texture, color, shape,
pattern, style, and material (we provided description examples of several texture
categories in the guidelines).

\para{Verification}
After collecting the raw annotations, we manually verified all of them and
removed annotations that were irrelevant.
For example, a breakfast waffle may have descriptions about the
related food items such as strawberries instead of the texture which is our main goal.
We also removed all images from ``freckled" and ``potholed" categories
because they are primarily of human faces or scenes of roads 
with few texture-related terms in their descriptions.
We also excluded images with fewer than $3$ valid descriptions.

\begin{figure}[t]
\centering
\makebox[0pt][c]{\parbox{\textwidth}{%
\begin{minipage}[t]{0.35\hsize}
    \begin{center}
    \resizebox{\linewidth}{!}{
    \begin{tabular}{l|c|c}
        \textbf{Statistics} & overall & frequent\\
        \hline
        \#images & 5369 & -\\
        \hline
        {\#phrases} & 22,435 & 655\\
        \hline
        {\#words} & 7681 & 1673\\
        \hline
        {\#descriptions per image} &  4.60 & -\\
        \hline
        {\#phrases per image} & 16.64 & 11.61 \\
        \hline
        {\#words per description} & 7.13 & 6.69 \\
        \hline
        {\#words per phrase} & 3.93 & 1.19 \\
    \end{tabular}}%
    \end{center}
\end{minipage}
\hfill
\begin{minipage}[t]{0.62\hsize}
    \begin{center}
    \resizebox{\linewidth}{!}{
    \includegraphics[width=\linewidth]{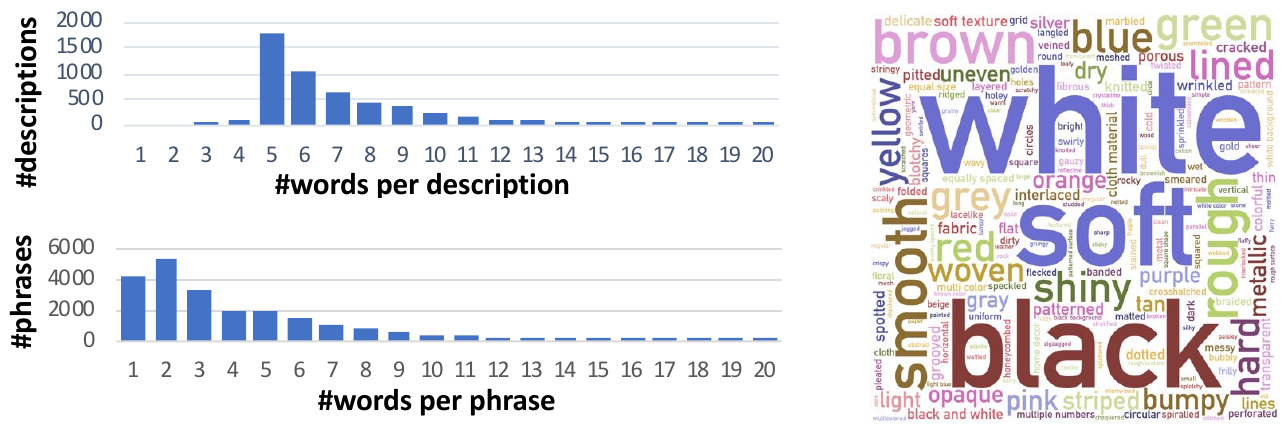}
    }
    \end{center}
\end{minipage}
}}
\caption{\textbf{Statistics of \dtdd.}
  The ``overall'' column in the table shows the statistics of all data, while
  the ``frequent" column only considers the phrases (or
words) that occur at least 10 (or 5) times in the training split which
forms our evaluation benchmark.
The cloud of phrases has the font sizes proportional to square-root of
frequencies in the dataset. The vocabulary significantly expands the 47
attributes of DTD.
}
\label{fig:stats}
\end{figure}

\para{Post-processing}
We found that the annotations (as seen in Figure~\ref{fig:data})
describing aspects of texture are often expressed as a set of
phrases separated by commas, instead of a fully grammatical sentence.
We did find some users who provided long unbroken sentences, but
these were few and far between.
Therefore, we represent each description as a set of
phrases indicated by commas (``,") or semicolons (``;'').
For example, the first description of the top-right image in Figure~\ref{fig:data} is: “spiralled, rounded, thick, light colour, rope type”, and it’s split into 5 phrases: “spiralled”, “rounded”, “thick”, “light colour”, “rope type”. 
For the purpose of evaluation, we consider words that appear at least
$5$ times and phrases that appear at least $10$ times in the training split of the dataset, 
which results in $655$ unique phrases.
Although some long descriptions are lost in the process
and most of the phrases are short (mostly within three words as seen in the lower histogram in Figure~\ref{fig:stats}),
the collection of phrases captures a rich set of describable attributes
for each image.
Modeling the space of phrases poses significant challenges to existing
techniques for language and vision (Section~\ref{sec:analysis}).

\subsection{Tasks and Evaluation Metrics}\label{sec:data-tasks}

The annotation for each image is in the form of a set of descriptions,
with each description in the form of a set of phrases. A phrase is an
ordered list of words.
We consider several tasks and evaluation metrics on this dataset
described next.

\para{Phrase retrieval}
Given an image, the goal is to rank phrases 
$p \in {\cal P}$ that are relevant to the image.
Here ${\cal P}$ is the set of all possible phrases, restricted to 655
frequent ones.
For each image, the set of ``true'' relevant phrases are obtained by
taking the union of phrases from all descriptions of the image.
We can evaluate the ranked list using various metrics described as follows:
\begin{itemize}
    \item {Mean Average Precision (MAP):} area under the
      precision-recall curve;
    \item {Mean Reciprocal Rank (MRR):} One over the ranking of first correct phrase;
    \item {Precision at K (P@K):} precision of the top K ranked phrases ($K\in \{5, 20\}$);
    \item {Recall at K (R@K):} recall of the top K ranked phrases ($K\in \{5, 20\}$).
\end{itemize}

\para{Image retrieval from a phrase}  The task is to retrieve images given a
query phrase.
When taking phrases as the query, we consider all phrases $p\in
{\cal P}$ as before and ask the retrieval model to rank all images in the
test or validation set.
The ``true'' list is all images that contain the
phrase (in any of its descriptions).
We consider the same metrics as the phrase retrieval task.

\para{Image retrieval from a description} When using descriptions
the query, we consider all description $s \in {\cal  
  S}$ as the input.
Here ${\cal S}$ is the set of all descriptions in the test or
validation set.
We ask the retrieval model to rank all images in the corresponding
set.
We evaluate the rank of the image from which the description was collected
(MRR metric).
This metric allows us to evaluate the compositional properties of texture
over phrases (\eg, ``red dots'' + ``white background'').
While we only quantitatively evaluate phrases and descriptions in the dataset, the ranking models can potentially
generalize to novel descriptions or phrases over the seen words.
We present qualitative results and a detailed study of the models in
Section~\ref{sec:exp} and~\ref{sec:analysis}.

\para{Description generation}
The task is to generate a description for an input image.
Given each image $I$, we compare the generated description against the
set of its collected descriptions $\{S_1, S_2, \ldots, S_k\}$ using standard 
metrics for image captioning including
BLEU-1,2,3,4~\cite{papineni2002bleu}, METEOR~\cite{banerjee-lavie-2005-meteor},
Rouge-L~\cite{rouge} and CIDEr~\cite{vedantam2015cider}.
However, we note that the task is open-ended and qualitative
visualizations are just as important as these metrics.

%% file: 4_method.tex
\section{Methods}
\label{sec:method}

We investigate three techniques to learn the mapping between
visual texture and natural languages on our dataset --- a discriminative approach, a metric learning approach, and a language generation approach.
They are explained in detail in the next three sections.

\subsection{A Discriminative Approach}
A simple baseline is to treat each phrase $p \in {\cal P}$ as a binary
attribute and train a multi-label classifier to map the images to
phrase labels.
Given a texture image $I$, let $\psi(I)$ be an embedding computed using
a deep network.
We use activations from layer 2 and layer 4 of
ResNet101~\cite{he2016deep} with mean-pooling over spatial locations
as the image embedding.
A comparison of features from different ResNet layers is included in the supplemental material.
For the classification task, we attach a classifier head $h$ to map the
embeddings to a 655-dimensional space corresponding to each phrase in our
frequent set ${\cal P}$.
The function $h$ is modeled as a two-layer network -- the first is
fully-connected layer with 512 units with BatchNorm and ReLU
activation; the second is a linear layer with 655 units followed by
sigmoid activation.
Given a training set of $\{(I_i, Y_i)\}_{i=1}^N$ where $Y_i$ is the
ground-truth binary labels across 655 classes for image $I_i$, the
model is trained to minimize the binary cross-entropy loss:
$L_{BCE}=\sum_{i} \ell_{bce}(h\circ \psi(I_i), Y_i)$,  where
$\ell_{bce}(y,z) = \sum_i \left(z_i\log(y_i) -
(1-z_i)\log(1-y_i)\right)$.

\para{Training details} 
The ResNet101 is initialized with weights pre-trained on ImageNet~\cite{imagenet_cvpr09} and fine-tuned on
our training data
for $75$ epochs using the
Adam optimizer~\cite{kingma2014adam} with an initial learning rate at
$0.0001$. We use image size 224$\times$224 for all our experiments. The hyper-parameters are selected on the validation set.

\para{Evaluation setup} 
The classification scores over each
phrase for each image are directly used to rank images or phrases
for phrase retrieval or image retrieval with phrase input.
Retrieving images given a description is more challenging since we
need to aggregate the scores corresponding to different phrases,
and the phrases in input descriptions may not be in ${\cal P}$.
We found the following strategy works well:
Given a description $S = \{P_1, P_2, \ldots, P_n\}$ and an image $I$, obtain the scores
for each phrase $s(P_i) = \sigma(h \circ \psi(I))_k$ where $k$ is the
index of the phrase $P_i \in {\cal P}$. If the phrase is not in the
set, we consider all its sub-sequences that are present in ${\cal P}$ and average the scores of them instead. For example, if the phrase ``red maroon dot'' is not
present in ${\cal P}$, we consider all sub-sequences \{red maroon,
maroon dot, red, maroon, dot\}, score each that is present in ${\cal
  P}$ separately and then average the scores.
By concatenating the top 5 phrases for an image we can also use the
classifier to generate a description for an image.
The key disadvantage of the classification baseline is that it treats
each phrase independently, and does not have a natural way to score
novel phrases 
(our baseline using sub-sequences is an attempt to
handle this).

\subsection{A Metric Learning Approach}
The metric learning approach aims to learn a common embedding over
the images and phrases such that nearby image and phrase pairs in the
embedding space are related.
We adopt the standard metric learning approach based on
triplet-loss~\cite{hoffer2015deep}.
Consider an embedding of an image $\psi(I)$ and of a phrase $\phi(P)$ in
$\mathbb{R}^d$. Denote $||\psi(I) - \phi(P)||_2^2$ as the squared
Euclidean distance between the two embeddings.
Given an annotation $(I, P)$ consisting of a positive (image, phrase) pair, we
sample from the training set a negative image $I'$ for $P$, and a negative phrase $P'$ for $I$.
We consider two losses; one from the negative phrase:
$$L_p(I, P, P')=\max(0, 1 + ||\psi(I) - \phi(P)||_2^2 - ||\psi(I) -
\phi(P')||_2^2)$$
and another from the negative image:
$$L_i(P, I, I')=\max(0, 1 + ||\psi(I) - \phi(P)||_2^2 - ||\psi(I') -
\phi(P)||_2^2)$$
The metric learning objective is to learn embeddings $\psi$ and $\phi$
that minimize the loss $L=\mathbb{E}_{(I,P), (I', P')} \left(L_p+L_i\right)$ over the
training set.

For embedding images, we use the same encoder as the
classification approach with features from layer 2 and 4 from ResNet101. 
We add an additional linear layer
with 256 units resulting in the embedding dimension 
$\psi(I) \in
\mathbb{R}^{256}$.
One advantage of the metric learning approach is that it allows us to
consider richer embedding models for phrases.
Specially we consider the following encoders:
\begin{itemize}
\item \textbf{Mean-pooling}:
$\phi_{mean}(P)=\frac{1}{N_w}\sum_{w\in \texttt{tokenize}(P)}\texttt{embed}(w)$, 
where $\texttt{tokenize}(\cdot)$ splits the phrase into a list of
words, $\texttt{embed}(\cdot)$ encodes each token into $\mathbb{R}^{300}$.

\item \textbf{LSTM}~\cite{schuster1997bidirectional}:
$\phi_{lstm}(P)=
\texttt{biLSTM}[\texttt{embed}(w) \text{ for } w \text{ in } \texttt{tokenize}(P)]$,
with the same $\texttt{tokenize}(\cdot)$ and  $\texttt{embed}(\cdot)$
as above. $\texttt{biLSTM}(\cdot)$ is a bi-directional LSTM with a single layer and hidden dimension 256 that returns the
concatenation of the outputs on the last token from both directions.

\item \textbf{ELMo}~\cite{Peters:2018}:
$\phi_{elmo}(P)=\texttt{ELMo}(P)$,
where $\texttt{ELMo}(\cdot)$ uses pre-trained ELMo model~\cite{elmo_wt}
with its own tokenizer, 
and outputs the average embedding of all tokens in the phrase $P$.

\item \textbf{BERT}~\cite{devlin2018bert}:
$\phi_{bert}(P)= 
\texttt{BERT}(P)$,
where $\texttt{BERT}(\cdot)$ uses pre-trained BERT model~\cite{bert_wt}
with its own tokenizer, 
and outputs the average of last hidden states of all tokens in the
phrase $P$.
\end{itemize}
To compute the final embedding of the phrase $\phi(P)$, we add a linear
layer to map the embeddings to 256 dimensions
compatible with the image embeddings.

\para{Training details}
We trained this model on our training split using the Adam
optimizer~\cite{kingma2014adam} with an initial learning rate at
$0.0001$. We found this model to be more prone to over-fitting than the classifier.
Stopping the training when the image retrieval and phrase retrieval MAP on the validation set stops improving was effective.
Same as the classifier, ResNet101 is initialized with ImageNet~\cite{imagenet_cvpr09} weights and fine-tuned on our data.
$\texttt{embed}(\cdot)$ in $\phi_{mean}$ and $\phi_{lstm}$ was initialized with FastText embeddings~\cite{bojanowski2017enriching, fasttext_emb} and tuned end-to-end.
Pre-trained encoders $\phi_{elmo}$ and $\phi_{bert}$ were fixed in our training.

\para{Evaluation setup}
Given the joint embedding space, one can retrieve phrases for each image
and images for each phrase based on the Euclidean distance.
Similar to the classifier we concatenate the top 5 retrieved phrases
as a baseline description generation model.
We also investigate a metric learning approach over descriptions
rather than phrases where the positive and negative triplets are
computed over (image, description) pairs.
The language embedding models are the same since they can handle
descriptions of arbitrary length.

\subsection{A Generative Language Approach}
We adopt the Show-Attend-Tell
model~\cite{xu2015show}, a widely used model for image captioning.
It combines a convolutional network to encode input images with an attention-based LSTM
decoder to generate descriptions.
Following the default setup, we encode images into
the spatial features from the 4-th layer of ResNet101 (initialized
with ImageNet~\cite{imagenet_cvpr09} weights).
The word embeddings are initialized from FastText~\cite{bojanowski2017enriching, fasttext_emb}. 
The entire model is then trained end-to-end on the training set, using the Adam
optimizer~\cite{kingma2014adam} with initial learning rate
$0.0001$ for the image encoder and $0.0004$ for the language decoder.
We apply early stopping based on the BLEU-4 score of generated descriptions on the validation images.

This model is primarily for the description generation task. 
For evaluation, we apply beam search with a beam-size of 5 to compute the best description.

%% file: 5_experiment.tex
\section{Experiments and Analysis}

\label{sec:exp}
\begin{table}[t]
\caption{Phrase retrieval and image retrieval on \dtdd. 
    Metric learning models are trained with phrase input.
    Among the language encoders BERT works the best.}
  \resizebox{\linewidth}{!}{
    \begin{tabular}{c|l||*{6}{c}|*{6}{c}}
    \hline
    \multicolumn{1}{r}{} & \multicolumn{1}{c||}{\textbf{Task:}} & \multicolumn{6}{c|}{\textbf{Phrase Retrieval}}  & \multicolumn{6}{c}{\textbf{Image Retrieval}} \\
    \hline
    Data Split & \multicolumn{1}{c||}{Model} & \multicolumn{1}{c}{MAP} & \multicolumn{1}{c}{MRR} & P@5 & \multicolumn{1}{c}{P@20} & \multicolumn{1}{c}{R@5} & \multicolumn{1}{c|}{R@20} & \multicolumn{1}{c}{MAP} & \multicolumn{1}{c}{MRR} & P@5 & \multicolumn{1}{c}{P@20} & \multicolumn{1}{c}{R@5} & \multicolumn{1}{c}{R@20} \\
    \hline\hline
    \multirow{4}[2]{*}{Validation} & MetricLearning: MeanPool & 18.80 & 48.66 & 23.13 & 16.20 & 11.52 & 31.54 & 7.19  & 16.18 & 7.60  & 6.56  & 3.36  & 11.44 \\
          & MetricLearning: biLSTM & 23.53 & 58.78 & 31.85 & 18.73 & 15.83 & 36.31 & 8.31  & 17.46 & 8.15  & 7.06  & 4.21  & 13.40 \\
          & MetricLearning: ELMo & 28.13 & 68.46 & 37.02 & 21.11 & 18.44 & 41.12 & 11.25 & 24.05 & 12.79 & 10.27 & 5.85  & 18.57 \\
          & \textbf{MetricLearning: BERT} & \textbf{31.68} & \textbf{72.59} & \textbf{40.67} & \textbf{22.96} & \textbf{20.23} & \textbf{44.50} & \textbf{15.22} & \textbf{31.39} & \textbf{16.27} & \textbf{12.56} & \textbf{9.07} & \textbf{25.69} \\
    \hline
    \multirow{2}[2]{*}{Test} & Classifier: Feat 2,4 & 27.12 & 61.28 & 33.50 & 21.71 & 16.07 & 41.48 & \textbf{14.75} & \textbf{33.94} & \textbf{18.75} & \textbf{16.02} & \textbf{6.47} & \textbf{19.32} \\
          & MetricLearning: BERT & \textbf{31.77} & \textbf{74.12} & \textbf{41.70} & \textbf{23.60} & \textbf{20.17} & \textbf{45.04} & 13.50 & 31.12 & 16.52 & 14.57 & 5.24  & 17.32 \\
    \hline
    \end{tabular}
    }%
    
  \label{tab:retrieve}%
\end{table}%


\begin{table}[t]
\begin{center}
\makebox[0pt][c]{\parbox{\textwidth}{%
\begin{minipage}[t]{0.3\hsize}
    \caption{Retrieving texture images with descriptions as input.}
    \resizebox{\linewidth}{!}{
    \begin{tabular}{l|c}
        \textbf{Model} & \textbf{MRR}\\
        \hline
        \hline
        Classifier & 12.40 \\
        \hline
        MetricLearning(phrase) & 12.92 \\
        \hline
        \textbf{MetricLearning(description)} & \textbf{13.95} \\
        \hline
    \end{tabular}}%
    \label{tab:retrieve_desc}%
\end{minipage}
\hfill
\begin{minipage}[t]{0.66\hsize}
    \caption{Description generation on textures. Synthesizing descriptions from phrases
      retrieved by the metric-learning based approach outperforms other baselines.}
    \resizebox{\linewidth}{!}{
    \begin{tabular}{l|*{7}{c}}
        \textbf{Model} & \textbf{Bleu-1} & \textbf{Bleu-2} & \textbf{Bleu-3} & \textbf{Bleu-4} & \textbf{METEOR} & \textbf{Rouge-L} & \textbf{CIDEr}\\
        \hline
        \hline
        Classifier: top~5 & 68.07 & 46.17 & 28.39 & 14.44 & 19.89 & 48.13 & 44.73 \\
        \hline
        \textbf{MetricLearning: top~5} & \textbf{72.99} & \textbf{53.69} & \textbf{34.97} & \textbf{19.39} & \textbf{21.81} & \textbf{49.70} & \textbf{47.34}\\
        \hline
        Show-Attend-Tell & 59.90 & 40.41 & 26.52 & 16.35 & 19.92 & 46.64 & 37.47 \\
        \hline
    \end{tabular}}%
    \label{tab:caption}%
\end{minipage}
}}
\end{center}
\end{table}


\begin{figure}[t]
\centering
\includegraphics[width=\linewidth]{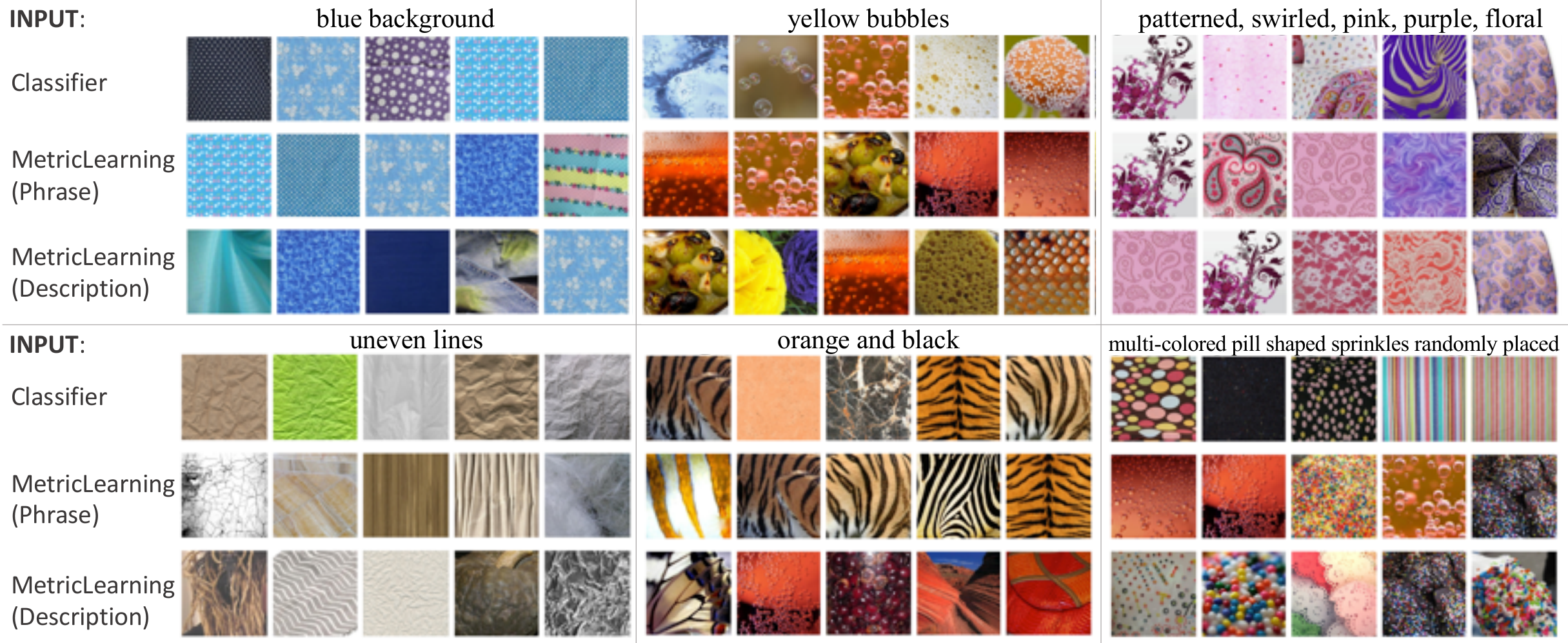}
\caption{\textbf{Retrieve \dtdd test images with language input.} 
We show top~5 retrieved images from the classifier and the metric learning model (trained with phrase or description input).
From left to right we show examples of 
(1) phrases the classifier has been trained on, 
(2) phrases beyond the frequent classes, and 
(3) full descriptions.
}
\label{fig:lang2img}
\end{figure}


\begin{figure}[t]
\centering
\includegraphics[width=\linewidth]{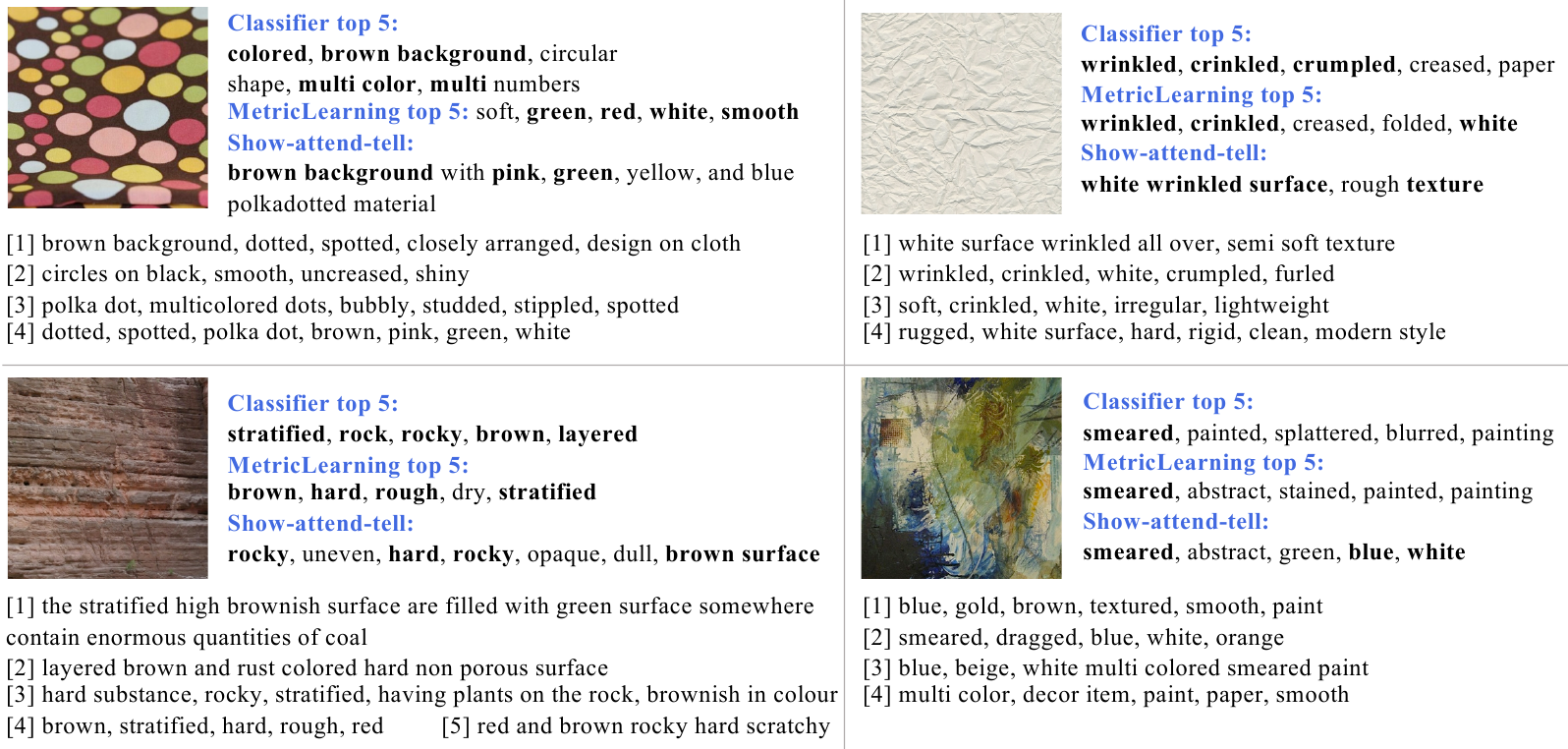}
\caption{\textbf{Phrase retrieval and description generation on \dtdd test images.}
For each input image, we list ground-truth descriptions beneath, and generated descriptions on the right. 
For the classifier and the metric learning model, we concatenate the top~5 retrieved phrases.
Bold words are the ones included in ground-truth descriptions.
}
\label{fig:img2lang}
\end{figure}

\subsection{Phrase and Image Retrieval}
\label{sec:exp-ret}

Table~\ref{tab:retrieve} and \ref{tab:retrieve_desc} compare the
classifier and the metric learning model on phrase and image retrieval tasks as described in
Section~\ref{sec:data-tasks}.
Figure~\ref{fig:lang2img} and~\ref{fig:img2lang} show examples of the
top 5 retrieved images and phrases.

In Table~\ref{tab:retrieve} we first compare language encoders on
the metric learning model.
The performance of both phrase and image retrieval depends largely on
the language encoder, and BERT performs the best.
The metric learning model is
better at phrase retrieval while the classifier is slightly better at
image retrieval.

Table~\ref{tab:retrieve_desc} shows results of image retrieval from
descriptions and here too the metric learning model outperforms the other two models.
As shown in Figure~\ref{fig:lang2img}-right,
although the models trained on phrases work reasonably well, the
metric learning model trained on descriptions handles long
queries better.

\subsection{Description Generation}
We compare the Show-Attend-Tell
model~\cite{xu2015show} with a retrieval based approach.
From the classifier and the metric learning model we retrieve the top 5
phrases and concatenate them in the order of their score to form a
description. As shown in Table~\ref{tab:caption}, the metric learning model
reaches higher scores on the metrics.
However, notice that in Figure~\ref{fig:img2lang} the
generative model's descriptions are more fluent and covers both the
color and pattern of the images, while the retrieval baselines
(especially the classifier) repeat phrases with similar meanings.

%% file: 6_analysis.tex
\subsection{A Critical Analysis of Language Modeling}
\label{sec:analysis}

\begin{figure}[t]
\centering
\includegraphics[width=\linewidth]{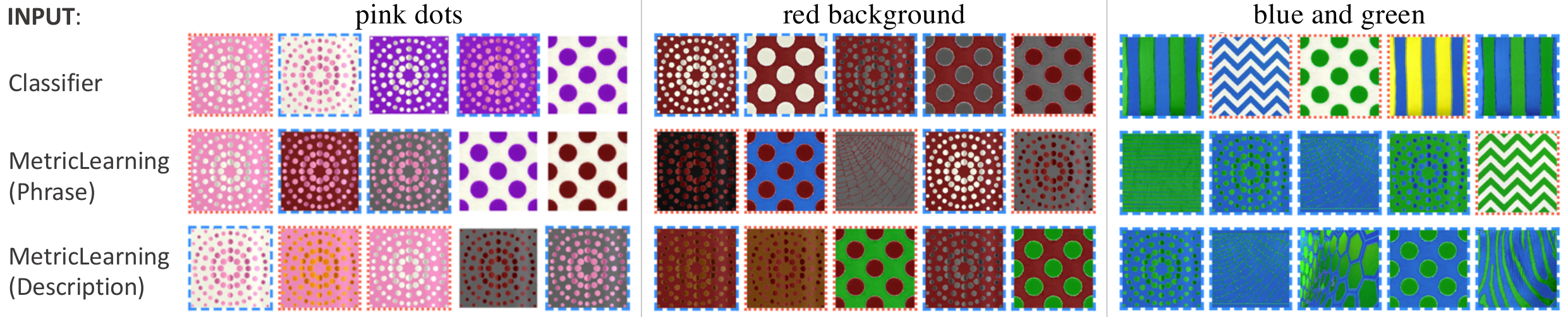}
\caption{\textbf{Retrieval on synthetic images.} Positive images are in dashed
  blue borders, hard negative ones are in dotted red borders.
}
\label{fig:synthetic}
\end{figure}

\begin{table}[t]
\caption{Image retrieval performance of R-Precision on synthetic tasks.}
\centering
\resizebox{0.85\linewidth}{!}{
    \begin{tabular}{l|cccc}
        \textbf{Model} & \textbf{~Foreground~} & \textbf{Background~} & \textbf{Color+Pattern~} & \textbf{Two-colors}\\
        \hline
        \hline
        Classifier & 45.45$\pm$20.34 & \textbf{59.82$\pm$9.63} & 35.95$\pm$21.48 & 26.82$\pm$14.17\\
        \textbf{MetricLearning - phrase}& 46.55$\pm$20.65 & 52.00$\pm$6.32 & \textbf{41.73$\pm$22.77} & \textbf{27.45$\pm$15.13}\\
        MetricLearning - description~ & \textbf{47.64$\pm$18.97} & 53.64$\pm$4.66 & 35.77$\pm$21.12 & 21.59$\pm$13.77\\
        \hline
        Random guess & 50.00 & 50.00 & 7.40 & 5.26\\
        \hline
\end{tabular}}
\label{tab:synthetic}
\end{table}

In this section, we evaluate the proposed models on tasks where we
systematically vary the distribution of underlying texture attributes.
This is relatively easy to do for textures than for natural images (\eg, changing the color
of dots) and allows us to understand the degree to which the models
learn disentangled representations.
We describe four tasks with varying degrees of difficulty to highlight
the strengths and weaknesses of these models.

\para{Automatically generating textures and their descriptions}
To systematically generate textures with descriptions,  we follow this procedure:
\begin{itemize}
\item Take the 11 most frequent colors in \dtdd (white, black, brown,
  green, blue, red, yellow, pink, orange, gray, purple) and set
  their RGB values manually.
\item Take 10 typical two-color images from ten different
  categories. We choose:
  \begin{itemize}
  \item Type A: 5 images with ``foreground on background'': [`dots', `polka-dots',  `swirls',`web', `lines' (thin lines on piece of paper)],
    and
    \item Type B: 5 images with no clear distinction between the
  foreground and background: [`squares' (checkered), `hexagon',
    `stripes' (zebra-like), `zigzagged', `banded' (bands with similar width)].
  \end{itemize}

\item For each of these 10 images, we manually extract masks for the foreground and
background(Type~A), or two foreground colors(Type~B).

\item For each of the 10 images, generate a new image by picking 2
  different colors from the 11 and modify pixel values of the two
  regions using the corresponding RGB value. This results in
  10$\times$11$\times$10=1,100 images.
\item For each synthetic image, we construct
  the ground-truth description with the template as ``[color1] [pattern], [color2]
  background"(such as ``pink dots, white background") for Type~A, and ``[color1]
  and [color2] [pattern]"(such as ``yellow and gray squares") for Type~B.
\end{itemize}

\noindent
\textbf{Experiment 1: Foreground.} On Type~A set we construct:
\begin{itemize}
  \item \textbf{Query:} A query of the form ``[color=c] [pattern=p]'' (\eg ``pink dots'').
  \item \textbf{Positive set:} [color=c]  [pattern=p] on randomly
    colored background (\eg ``pink dots, white background'').
  \item \textbf{Negative set:} Randomly colored ($\neq$ c)
      [pattern=p] on [color=c] background (\eg ``blue dots, pink
      background'').
  \item \textbf{Result:} 
  Input the query description, we use the models to rank images from
  both the positive and negative set, and report R-Precision: the
  precision of top $R$ predictions, where $R$ is the number of
  positive images. The results are listed in
    Table~\ref{tab:synthetic} first column.
    Since half the images have
    the right attribute the chance performance is
    50\% and the various models are nearly at the chance level.
    Figure~\ref{fig:synthetic} shows that the model is unable to
    distinguish between ``pink dots'' and ``dots on a pink
    background''. This illustrates that the models are unable to associate color correctly with the foreground shapes.
\end{itemize}

\noindent
\textbf{Experiment 2: Background.} This is similar to Experiment 1 but
we focus on the background instead. On Type~A set we construct:
we know the name of its pattern (such
  as ``dots", ``squares", selected from the more frequent phrases that
  matches the category) and names of two colors (color1 and color2).
  \begin{itemize}
  \item \textbf{Query:} A query ``[color=c] background'' (\eg ``pink background'').
  \item \textbf{Positive set:} Randomly colored pattern on [color=c]
    background (\eg ``red dots on pink background'').
  \item \textbf{Negative set:} Random pattern of [color=c] on any
    [color$\neq$c] background (\eg ``pink dots on white
    background''). 
  \item \textbf{Result:} R-precision is shown in
    Table~\ref{tab:synthetic} second column.
    Once again the chance performance is
    50\% and the various models are nearly at the chance level.
    Figure~\ref{fig:synthetic}-middle shows that the model is unable to
    distinguish between ``red background'' and ``red dots on random
    background''.
\end{itemize}

\noindent
\textbf{Experiment 3: Color+Pattern.} On both Type~A and~B images we construct:
\begin{itemize}
  \item \textbf{Query:} A query ``[color=c] [pattern=p]'' (\eg ``pink dots'').
  \item \textbf{Positive set:} [color=c] [pattern=p] on random colored
    background, or with another color (\eg ``pink dots, white background'', ``pink and blue squares'').
  \item \textbf{Negative set:} [color=c] [pattern$\neq$p] or
    [color$\neq$c] [pattern=p]. The negative set
    contains images with the correct pattern but wrong color or the
    wrong pattern with the right color (\eg, ``red dots'' or ``pink stripes'').
    Similar patterns (\eg, ``lines'' \vs ``banded'') are not considered negative.
  \item \textbf{Result:} The positive and negative set is unbalanced
    which results in a chance performance of 7.4\%. The models
    presented in the earlier section are able to rank the correct
    color and pattern combinations ahead of the negative set and
    achieve a considerably higher performance.    
\end{itemize}

\noindent
\textbf{Experiment 4: Two colors.}  On both Type~A and~B images we construct:
\begin{itemize}
  \item \textbf{Query:} A query ``[color=c1] and [color=c2]'' (\eg ``pink
    and green'').
  \item \textbf{Positive set:} [color=c1] of random pattern on [color=c2] background (\eg ``pink dots on green background''), or [color=c1] and [color=c2] of random pattern (\eg ``pink and green squares'').
  \item \textbf{Negative set:} pattern with one color from
    \{c1, c2\} and another color $\neq$\{c1,c2\} (\eg, ``pink
    dots on yellow background", ``green and blue stripes").
  \item \textbf{Result:} The positive and negative set are unbalanced
    which results in a chance performance of 5.26\%. The models
    once again are able to rank the two color combinations ahead of the negative set and
    achieve a considerably higher
    performance. Figure~\ref{fig:synthetic}-right shows an example.
\end{itemize}

\para{Summary} 
These experiments reveal that these models do indeed exhibit some high-level discriminative abilities (Exp. 3, 4), but they fail to disentangle properties such as the color of the foreground elements
from background (Exp. 1, 2). This leaves much room for improvement,
motivating future work, such as those that enforces spatial agreement between
different attributes.

%% file: 7_application.tex
\section{Applications}
\label{sec:app}
\begin{figure*}[t]
    \centering
    \includegraphics[width=\textwidth]{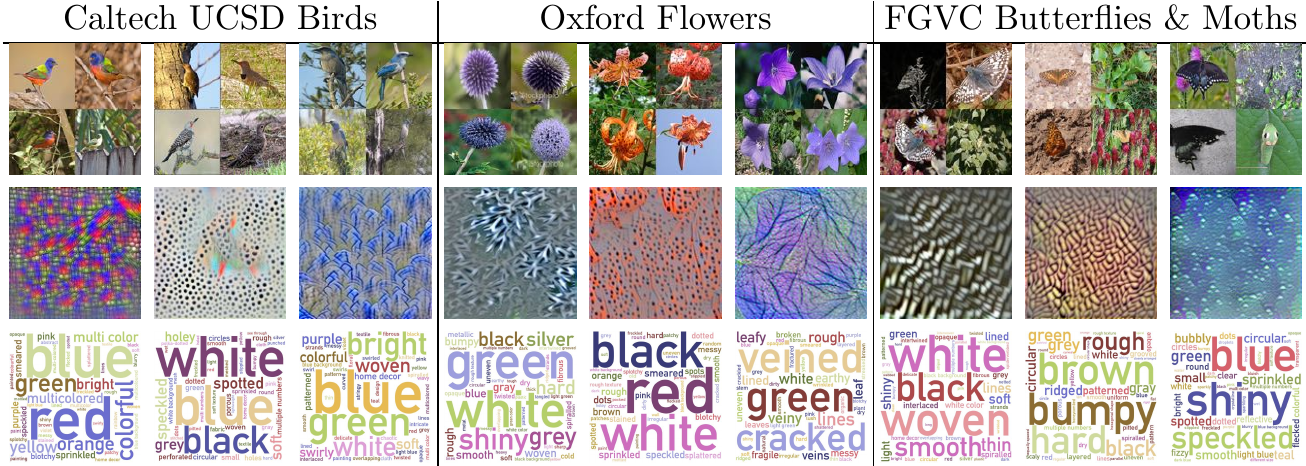}
    \caption{
      Fine-grained categories visualized as their training
      images (top row), maximal texture images (middle row), and
      texture attributes (bottom row). The size of each phrase in the
      cloud is inversely decided by its Euclidean distance to the
      input maximal texture image calculated by the triplet model.}
    \label{fig:fgvc}
\end{figure*}

\para{Describing textures of fine-grained categories}
We analyze how the categories in fine-grained domains can be described by their texture.
We consider categories from Caltech-UCSD Birds (CUB)~\cite{WahCUB_200_2011}, Oxford
Flowers~\cite{Nilsback08}, and FGVC Butterflies and
Moths~\cite{fgvcbutterlies} datasets.
For each category, we follow the visualizing deep
texture representations~\cite{lin2016visualizing} to generate
``maximal textures'' --  inputs that maximize the class probability
using multi-layer bilinear
CNN classifier~\cite{lin2018bilinear}.
These are provided as input to our metric learning model (with BERT encoder
and phrase input) trained on \dtdd to retrieve the top phrases.
Figure~\ref{fig:fgvc} shows several categories with their maximal
textures and a ``phrase cloud'' of the top retrieved phrases.
These provide a qualitative description of each category.

\begin{figure}[t]
\centering
\includegraphics[width=\textwidth]{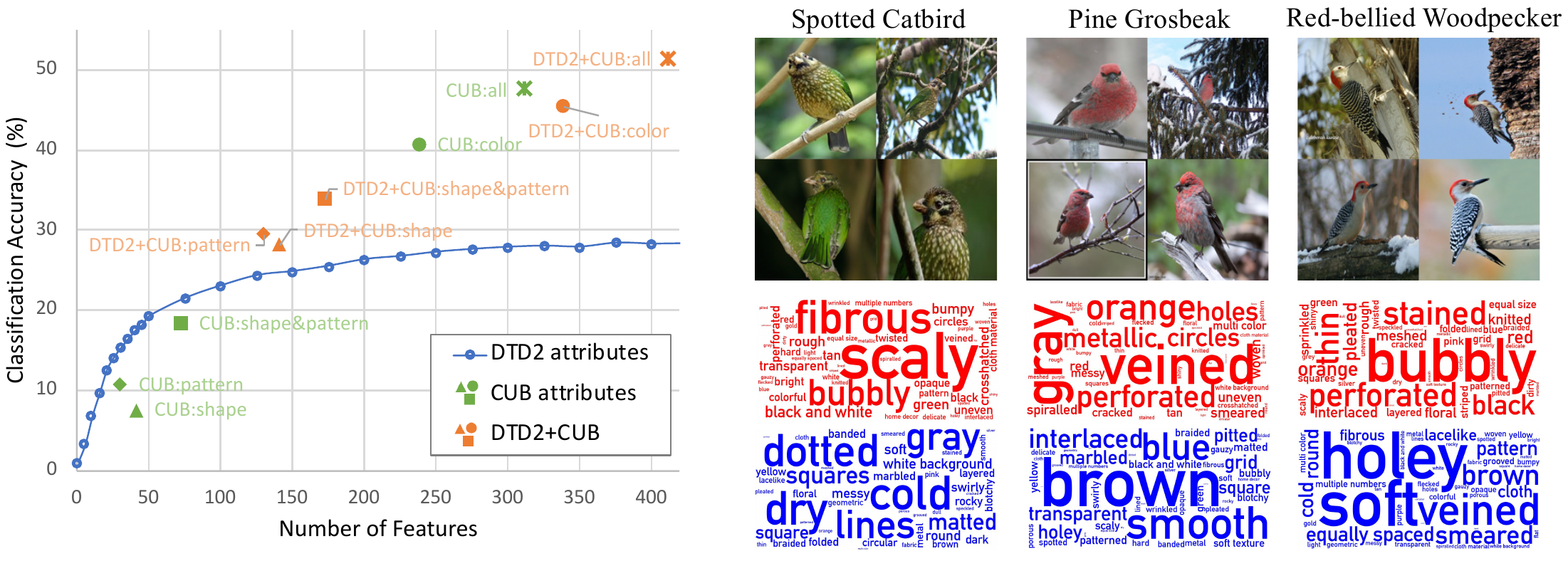}
\caption{\textbf{Classification on CUB dataset with \dtdd texture attributes. }
\emph{Left:} classification accuracy \vs number of input features. 
Orange and green markers with the same shape are comparable with the same set of CUB attributes with or without the \dtdd attributes.
\emph{Right:} The phrase clouds display important phrases for a few bird categories. 
Red phrases correspond to positive weights and blue are negative for a linear classifier for the category. Font sizes represent the absolute value of the coefficient.
}
\label{fig:cub}
\end{figure}

\para{Fine-grained classification with texture attributes}
Here we apply models trained on our \dtdd on the CUB dataset to show that
embedding images into the space of texture attributes allows interpretable models for discriminative classification.
Specifically, we input each image from the dataset to our
phrase classifier (trained on \dtdd and fixed) and obtain the 
log-likelihood over the $655$ texture phrases as an embedding.
We train a logistic regression model for the 200-way classification task.
The dataset also comes with $312$ binary attributes that 
describe the shape, pattern and color of specific parts of a bird,
such as ``has tail shape squared tail", ``has breast pattern spotted",
``has wing color yellow"
(42 attributes for ``shape", 31 for
``pattern" and 239 for ``color").
We also train a logistic regression classifier on top of these attributes.

Figure~\ref{fig:cub} shows the performance by varying the number of
texture phrases ranked by their frequency on \dtdd as the blue curve.
It also shows a comparison of bird-specific
attributes from CUB with generic texture attributes learned from \dtdd.
Results using CUB attributes are shown in green, while
those using combinations of CUB and texture attributes are shown in orange.
Texture attributes are able to distinguish bird species with an accuracy of $28.5\%$, outperforming CUB shape and pattern
attributes.
However, they do not outperform the part-based color attributes that
are highly effective.
Yet, combining CUB attributes with texture attributes lead to
consistent improvements.
On the right is the visualization of discriminative texture attributes for some categories: we display phrases with the most positive weights in red, and those with the most negative weights in blue.
They provide a basis for interpretable explanations of discriminative features without requiring a category-specific vocabulary.

%% file: 8_conclusion.tex
\section{Conclusion}
We presented a novel dataset of textures with natural language descriptions and analyzed the performance of several language and vision models.
The domain of texture is poses challenges to existing models which fail to learn a sufficiently disentangled representation leading to poor generalization on synthetic tasks.
Yet, the models show some generalization to novel domains and enabling us to provide interpretable models for describing some fine-grained domains. 
In particular they are complementary to existing domain-specific attributes on the CUB dataset.

\textbf{Acknowledgements.} We would like to thank Mohit Iyyer for helpful discussions and feedback. The project is supported in part by NSF grants \#1749833 and \#1617917. Our experiments were performed in the UMass GPU cluster obtained under the Collaborative Fund managed by the Mass. Technology Collaborative.

%% file: supp/supp.tex
\section*{Supplementary Material}
\setcounter{section}{0}
\setcounter{figure}{0}
\setcounter{table}{0}

\section{Dataset statistics and visualizations}
\begin{figure}[h]
\centering
\includegraphics[width=\linewidth]{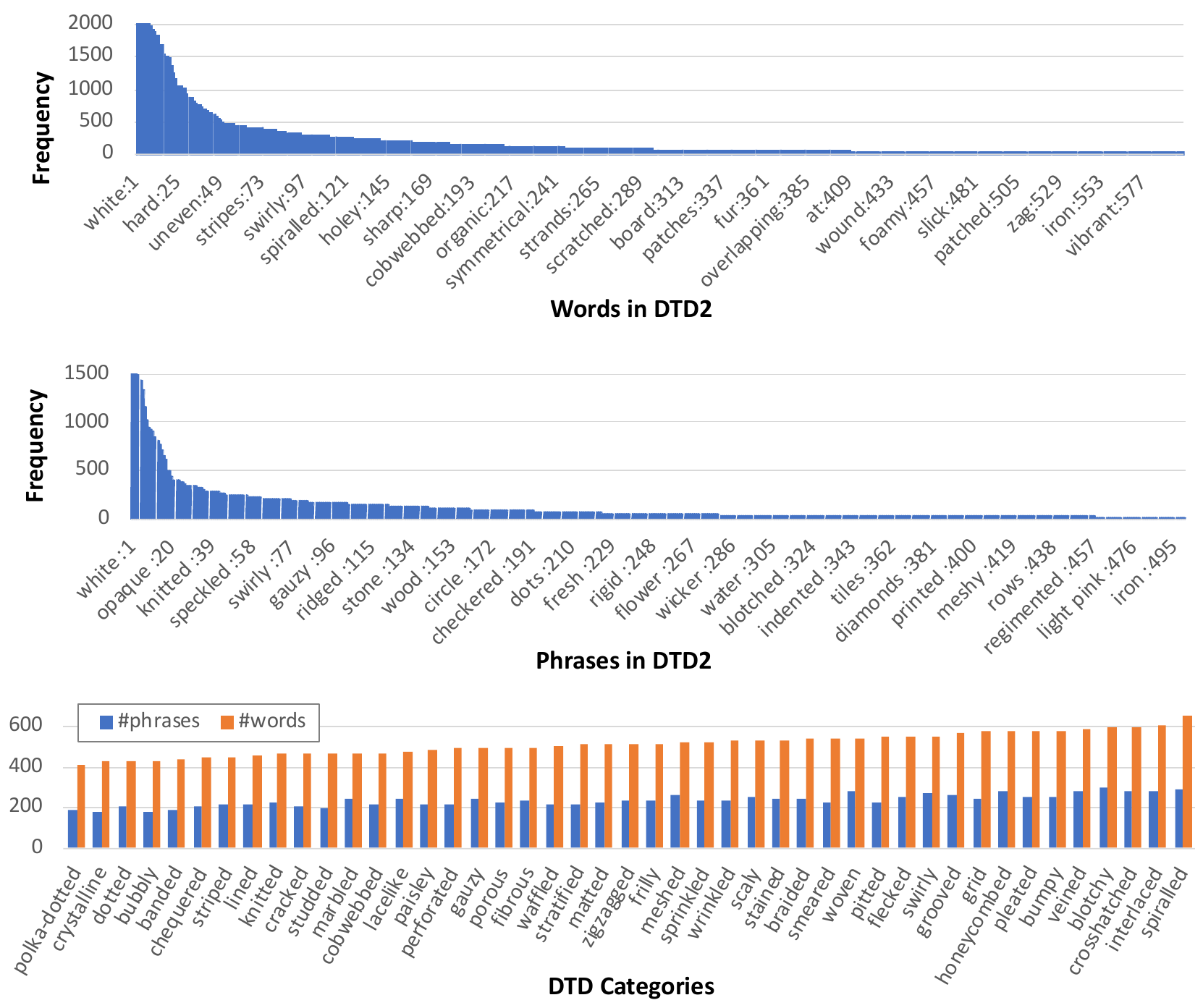}
\caption{\textbf{More Statics of \dtdd.}
On the top (middle) we show the histogram of word(phrase) frequency, where the x labels display some of the words(phrases) with their rankings in the dataset. 
On the bottom we show the number of unique words and phrases used to describe each DTD category, where we only count frequent words(phrases) that occur at least 5(10) times in training subset.}
\label{fig:stat}
\end{figure}

\begin{figure}
\centering
\includegraphics[width=\linewidth]{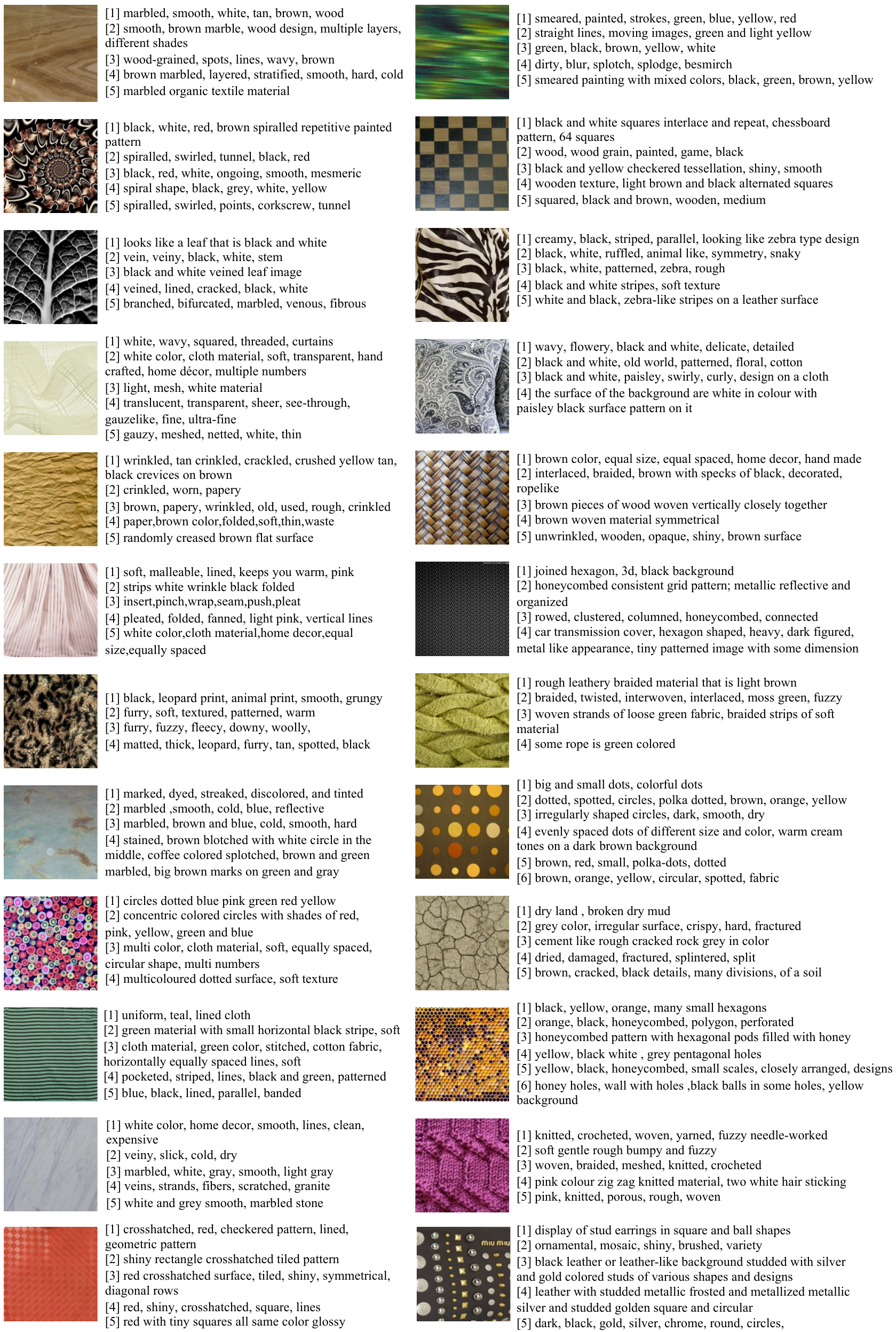}
\caption{More examples from \dtdd.}
\label{fig:data1}
\end{figure}

\begin{figure}
\centering
\includegraphics[width=\linewidth]{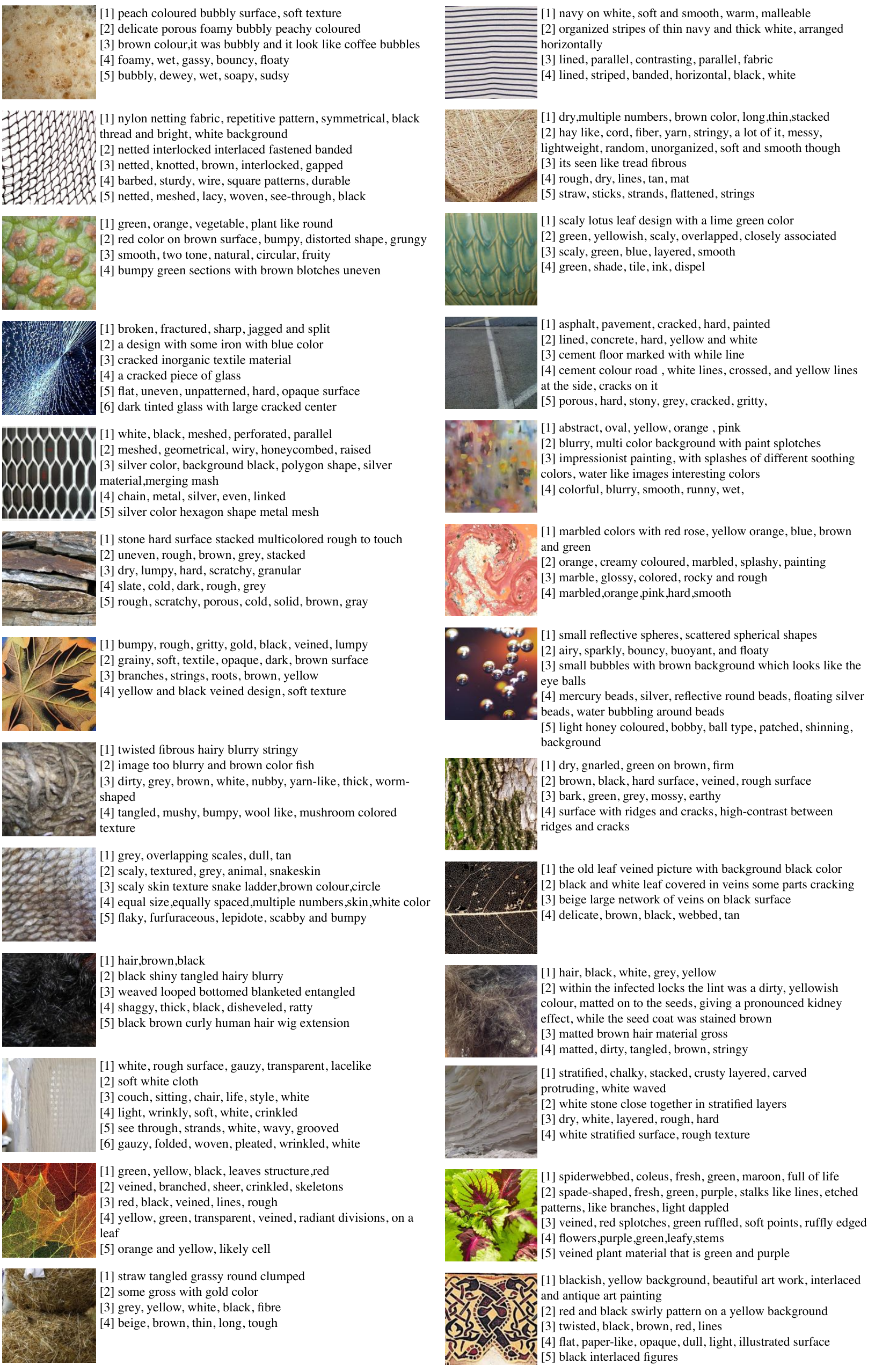}
\caption{More examples from \dtdd.}
\label{fig:data2}
\end{figure}

\begin{figure}
\centering
\includegraphics[width=\linewidth]{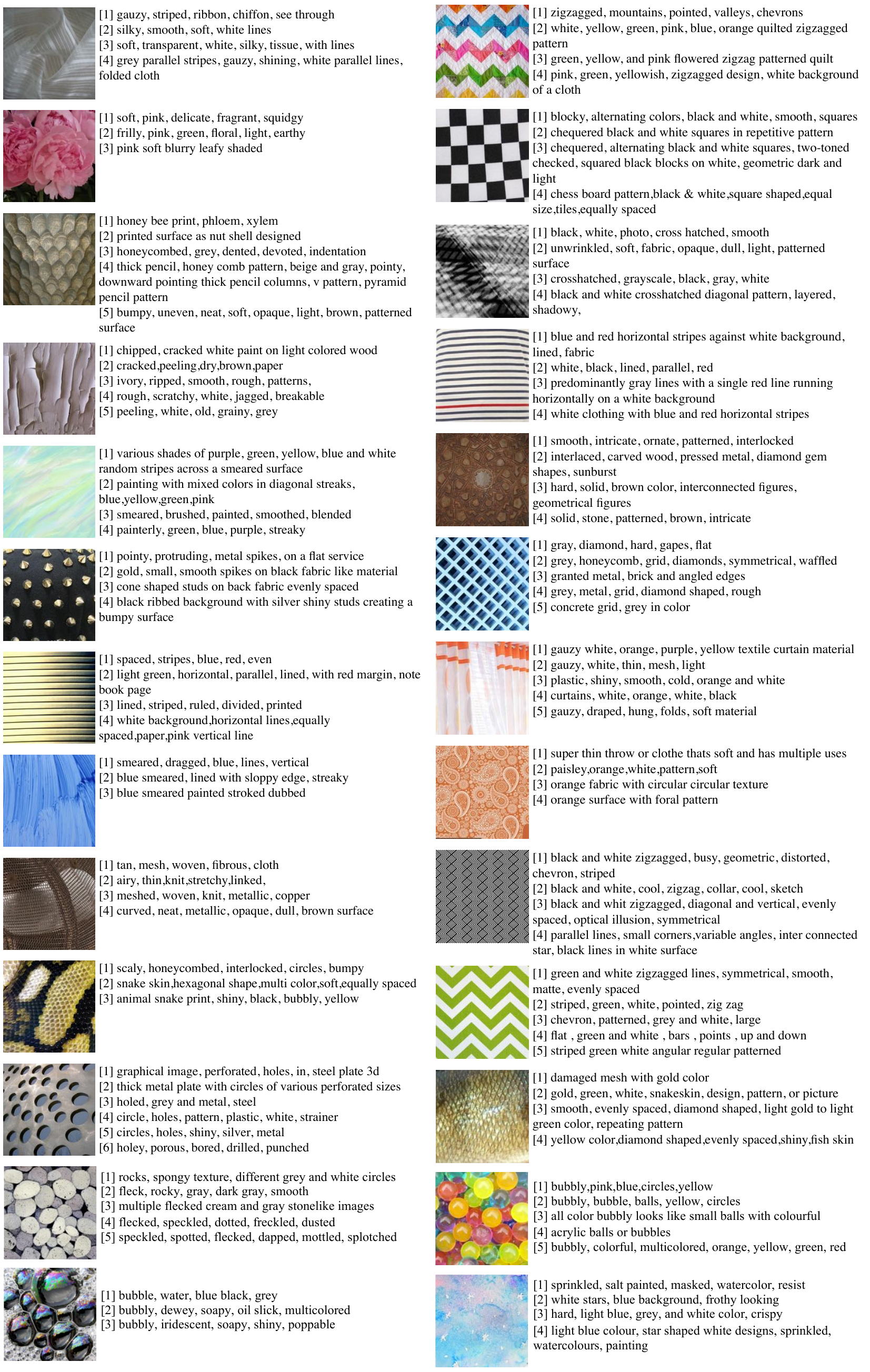}
\caption{More examples from \dtdd.}
\label{fig:data3}
\end{figure}

We provide further details about the \dtdd. 
Figure~\ref{fig:stat} shows the long-trail distribution of words and phrases in the dataset.
Descriptions of categories such as ``dots", ``lines", ``crystalline" and ``checkers" are simpler than those of ``spiralled'', ``interlaced'' and ``crosshatched'' as indicated by the number of unique words and phrases for each category.
More examples of images and annotations from the dataset are shown in  Figure~\ref{fig:data1},~\ref{fig:data2}, and~\ref{fig:data3}. 

\section{Comparison of ResNet101 features from different layers}
We compare features from layer-block 1 to 4 of ResNet101 in the classifier model, as shown in  Table~\ref{tab:retrieve_supp}.
Higher layer features perform better for phrase retrieval.
For image retrieval, better performance is achieved with the
combination of features from different layers.
We select to use the features from layer 2 and 4 for all classifiers
and metric learning models in all experiments with the classifier and the metric learning model.

\begin{table}[t]
\caption{Performance on phrase retrieval and image retrieval on \dtdd. 
  ``Classifier: Feat $x$" stands for the classifier with image
    features from ResNet layer block $x$ (or a concatenation of two layers.)}
  \resizebox{\linewidth}{!}{
    \begin{tabular}{c|l||*{6}{c}|*{6}{c}}
    \hline
    \multicolumn{1}{r}{} & \multicolumn{1}{c||}{\textbf{Task:}} & \multicolumn{6}{c|}{\textbf{Phrase Retrieval}}  & \multicolumn{6}{c}{\textbf{Image Retrieval}} \\
    \hline
    Data Split & \multicolumn{1}{c||}{Model} & \multicolumn{1}{c}{MAP} & \multicolumn{1}{c}{MRR} & P@5 & \multicolumn{1}{c}{P@20} & \multicolumn{1}{c}{R@5} & \multicolumn{1}{c|}{R@20} & \multicolumn{1}{c}{MAP} & \multicolumn{1}{c}{MRR} & P@5 & \multicolumn{1}{c}{P@20} & \multicolumn{1}{c}{R@5} & \multicolumn{1}{c}{R@20} \\
    \hline\hline
    \multirow{7}[1]{*}{Validation} & Classifier: Feat 1 & 13.10 & 37.20 & 16.05 & 10.68 & 4.94  & 13.04 & 10.64 & 25.06 & 11.57 & 9.37  & 6.13  & 17.78 \\
          & Classifier: Feat 2 & 17.65 & 44.91 & 22.41 & 14.59 & 6.85  & 17.60 & 13.00 & 29.24 & 14.60 & 11.08 & 8.52  & 22.54 \\
          & Classifier: Feat 3 & 26.43 & \textbf{60.52} & 32.47 & 20.71 & 9.93  & 25.00 & 15.62 & 31.79 & 17.28 & 13.34 & 9.42  & 28.52 \\
          & Classifier: Feat 4 & 26.51 & 59.24 & \textbf{33.07} & 20.84 & \textbf{10.07} & 25.16 & 15.85 & \textbf{33.06} & 17.83 & 13.02 & 9.94  & 27.28 \\
          & Classifier: Feat 1,4 & 25.78 & 58.28 & 31.58 & 20.31 & 9.55  & 24.44 & 15.85 & 32.35 & \textbf{18.35} & 13.51 & 10.24 & 28.03 \\
          & \textbf{Classifier: Feat 2,4} & 26.57 & 59.19 & 32.65 & 21.11 & 9.99  & 25.50 & \textbf{16.19} & 32.53 & 17.47 & \textbf{13.56} & \textbf{10.63} & \textbf{28.69} \\
          &Classifier: Feat 3,4 & \textbf{26.66} & 60.38 & 32.20 & \textbf{21.22} & 9.81  & \textbf{25.68} & 16.04 & 31.18 & 17.59 & 13.50 & 10.33 & 28.32 \\
    \hline
    \multirow{2}[2]{*}{Test} & Classifier: Feat 2,4 & 27.12 & 61.28 & 33.50 & 21.71 & 16.07 & 41.48 & \textbf{14.75} & \textbf{33.94} & \textbf{18.75} & \textbf{16.02} & \textbf{6.47} & \textbf{19.32} \\
          & MetricLearning: BERT & \textbf{31.77} & \textbf{74.12} & \textbf{41.70} & \textbf{23.60} & \textbf{20.17} & \textbf{45.04} & 13.50 & 31.12 & 16.52 & 14.57 & 5.24  & 17.32 \\
    \hline
    \end{tabular}
    }%
    
  \label{tab:retrieve_supp}%
\end{table}%

\section{Performance on different categories and phrases}
\begin{figure}
\centering
\includegraphics[width=0.9\linewidth]{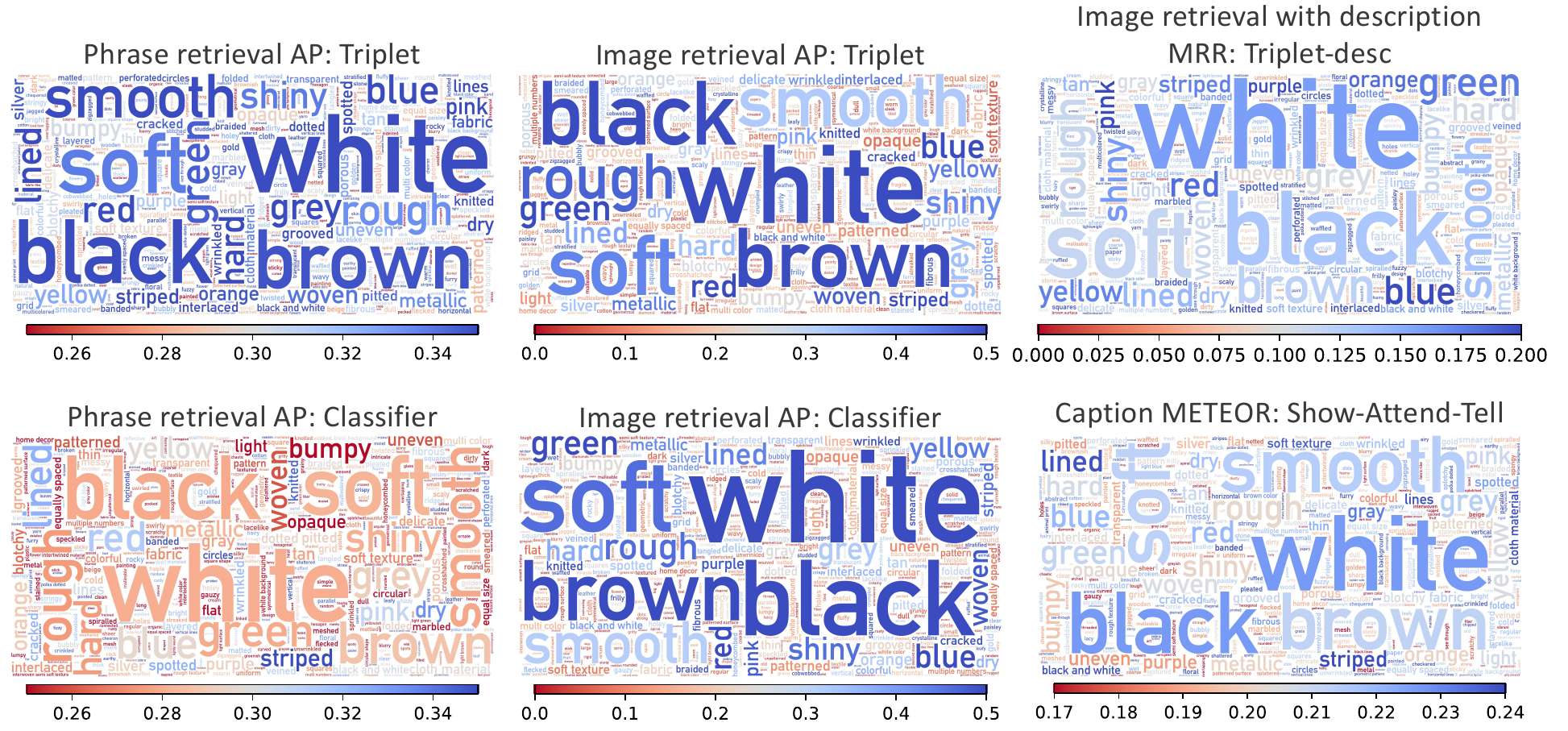}
\caption{\textbf{\dtdd test set performance per phrase on each task with selected models.}
Color of phrases represents the metric performance: blue is better and red is worse, as indicated in the color bars. Font sizes are proportional to square root of phrase frequencies.}
\label{fig:ph}
\end{figure}

\begin{figure}[t]
\centering
\includegraphics[width=0.95\linewidth]{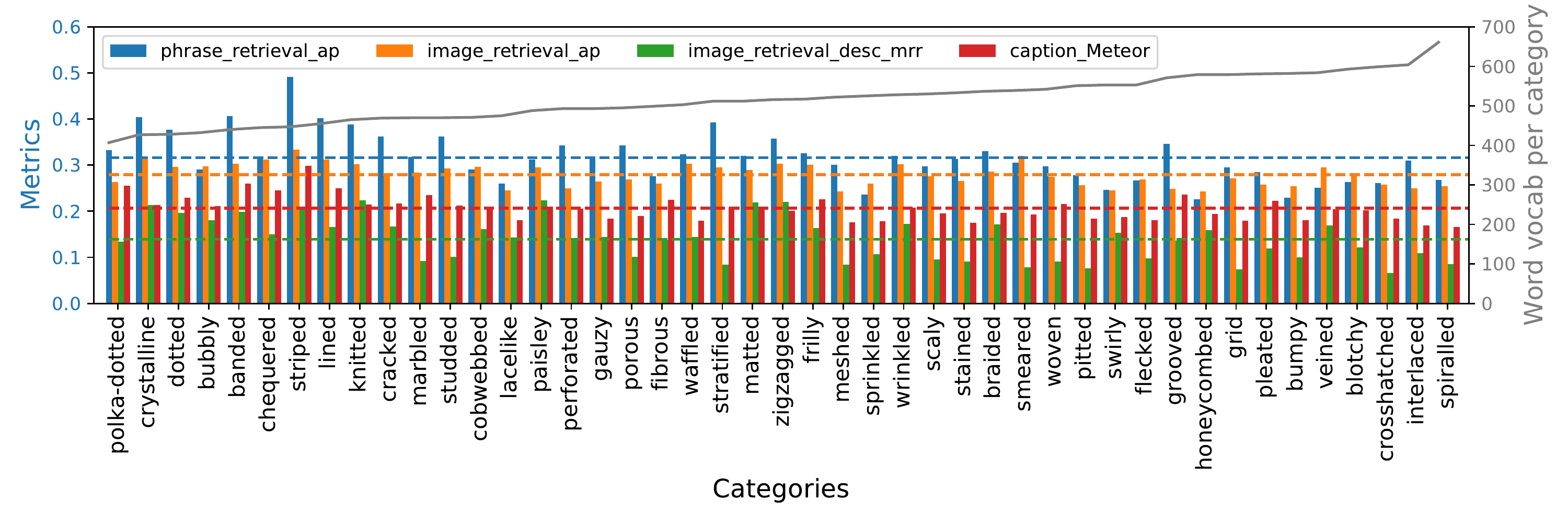}
\caption{\textbf{\dtdd test set performance per category on each task with our best model.}
Phrase retrieval: AP on our metric learning model; 
Image retrieval with phrase input: AP on our classifier;
Image retrieval with description input: MRR on our metric learning model(description input);
Captioning: METEOR on Show-Attend-Tell model.}
\label{fig:cat}
\end{figure}

Here we take a closer look at the performance of our models on different tasks on different phrases (Figure~\ref{fig:ph}) and texture categories (Figure~\ref{fig:cat}). The texture category is the category of the image in \dtd.
As seen in Figure~\ref{fig:ph}-middle, the performance is correlated with the frequency of phrases, especially for image retrieval.
As seen in Figure~\ref{fig:cat}, we have better phrase retrieval performance on simpler categories that come with smaller vocabulary size.

\section{Generated textures for analysis}
\begin{figure}[t]
\centering
\includegraphics[width=\linewidth]{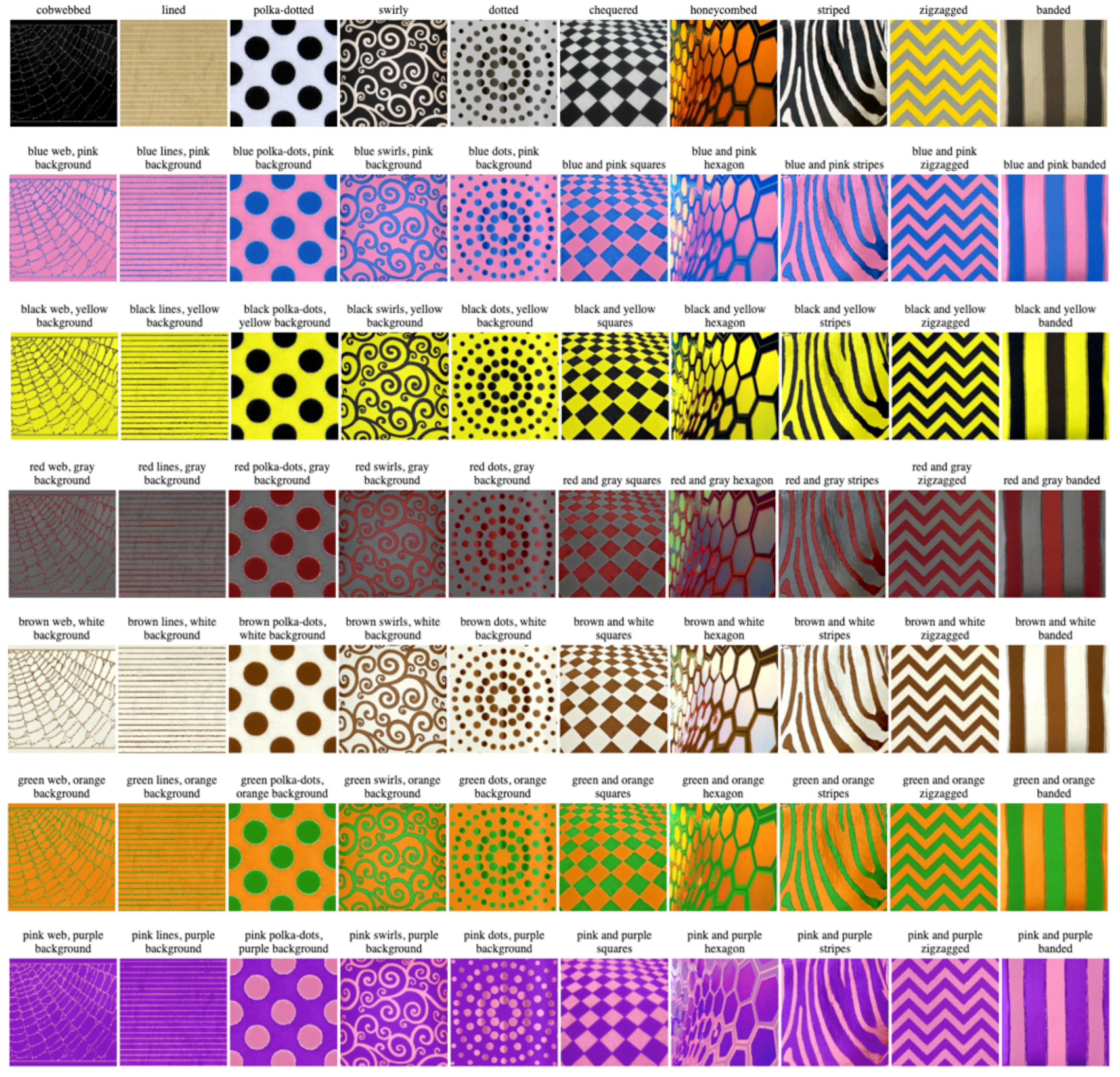}
\caption{\textbf{Original and generated texture images.}
The first row shows original images and their \dtd categories.
The rest rows are generated descriptions and images by sampling and modifying colors. The left 5 columns are images of Type A (foreground and background); the right 5 are Type B (no obvious foreground/background distinction). 
}
\label{fig:syn}
\end{figure}

Figure~\ref{fig:syn} shows the synthetically generated images and descriptions used in the experiments in Section 5.3.
Each image is manually separated into two color regions which are systematically varied. 
In each row of the figure we systematically vary different attributes such as the foreground attribute, foreground and background color, etc.
This is a simple yet effective way to generate natural textures.
Note how the texture properties are still maintained in each image, such as shading and winkles of zebra stripes.
This is much more challenging to do the same with natural images of scenes and objects.

